\documentclass[sigconf]{acmart}

\AtBeginDocument{%
  \providecommand\BibTeX{{%
    \normalfont B\kern-0.5em{\scshape i\kern-0.25em b}\kern-0.8em\TeX}}}

\setcopyright{rightsretained} 
\copyrightyear{2019}
\acmYear{2019}
\acmConference[SA '19 XR]{SIGGRAPH Asia 2019 XR}{November 17--20, 2019}{Brisbane, QLD, Australia}
\acmBooktitle{SIGGRAPH Asia 2019 XR (SA '19 XR), November 17--20, 2019, Brisbane, QLD, Australia}
\acmPrice{15.00}
\acmDOI{10.1145/3355355.3361896}
\acmISBN{978-1-4503-6947-3/19/11}

\begin{document}

\title{TouchVR: a Wearable Haptic Interface for VR Aimed at Delivering Multi-modal Stimuli at the User's Palm}

\author{Daria Trinitatova}
\affiliation{%
  \institution{Skolkovo Institute of Science and Technology (Skoltech)}
  \streetaddress{Nobelya Ulitsa 3}
  \city{Moscow}
  \country{Russia}
  \postcode{121205}}
  \email{daria.trinitatova@skoltech.ru}

\author{Dzmitry Tsetserukou}
\affiliation{%
  \institution{Skolkovo Institute of Science and Technology (Skoltech)}
  \streetaddress{Nobelya Ulitsa 3}
  \city{Moscow}
  \country{Russia}
  \postcode{121205}
}
\email{d.tsetserukou@skoltech.ru}

\renewcommand{\shortauthors}{Trinitatova and Tsetserukou}

\begin{abstract}
 TouchVR is a novel wearable haptic interface which can deliver multimodal tactile stimuli on the palm by DeltaTouch haptic display and vibrotactile feedback on the fingertips by vibration motors for the Virtual Reality (VR) user. DeltaTouch display is capable of generating 3D force vector at the contact point and presenting multimodal tactile sensation of weight, slippage, encounter, softness, and texture. The VR system consists of HTC Vive Pro base stations and head-mounted display (HMD), and Leap Motion controller for tracking the user's hands motion in VR. The MatrixTouch, BallFeel, and RoboX applications have been developed to demonstrate the capabilities of the proposed technology. A novel haptic interface can potentially bring a new level of immersion of the user in VR and make it more interactive and tangible. 
\end{abstract}

\ccsdesc[500]{Human-centered computing~Haptic devices}
\ccsdesc[300]{Human-centered computing~Virtual reality}
\ccsdesc{Computer systems organization~Robotics}

\keywords{Haptic Device, Tactile Display, Inverted Delta Robot, Multimodal stimuli, Virtual Reality}

\begin{teaserfigure}
  \includegraphics[width=\textwidth]{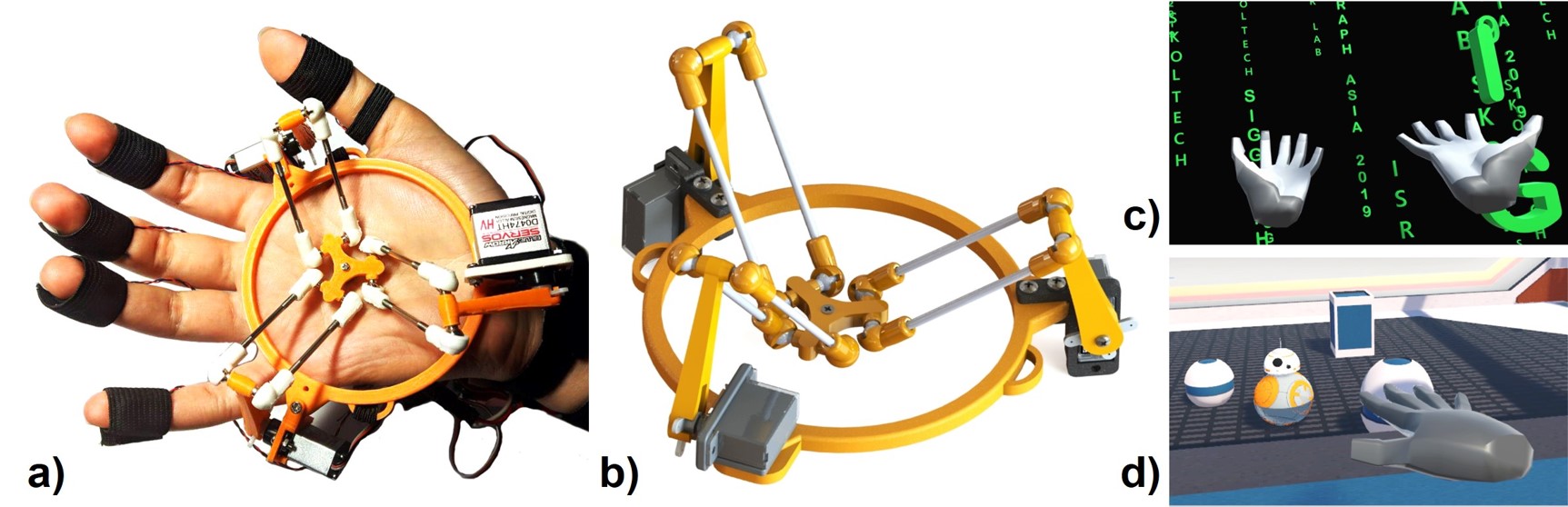}
  \caption{a) Novel wearable haptic interface TouchVR. b) A CAD model of haptic display. c) MatrixTouch application. d) RoboX application. }
  \Description{Enjoying the baseball game from the third-base
  seats. Ichiro Suzuki preparing to bat.}
  \label{fig:teaser}
\end{teaserfigure}

\maketitle

\section{Introduction}

Nowadays, many different entertainment applications, games, and simulators that implemented with VR/AR technologies are available in the market. However, the most VR/AR applications have a lack of providing feasible feedback to the user. In order to sense the physical interaction with the virtual environment, it is needed to reproduce tactile senses, which are a crucial component to accomplish the full immersion. Achieving a real touch simulation is a rather hard task which includes a representation of parameters of an object such as shape, hardness, weight, texture, and temperature.
Some wearable interfaces for simulating the grasping of rigid objects in a virtual environment have been proposed, such as Wolverine \cite{Choi2016}, and Grabity \cite{Choi2017}. They can provide the sensation of contact, gripping, gravity, and inertia.  A number of haptic interfaces provide both kinesthetic and cutaneous stimuli to the hand \cite{Son:2018:HFP:3242587.3242656}, \cite{Hinchet:2018:DWH:3242587.3242657}.
\par  We propose a novel wearable haptic interface for providing cutaneous feedback both at the fingertips and on a palm in VR environment (Fig. \ref{fig:teaser} (a)).

\section{System Description}
 A wearable haptic interface provides cutaneous feedback on the palm by DeltaTouch haptic display \cite{8816136} and vibrotactile feedback on the fingertips by vibration motors (tablet type, 8x3 mm). DeltaTouch is a wearable haptic display to provide multimodal tactile stimuli at the palm of the user based on the parallel inverted delta structure attached at the hand of the user (Fig. \ref{fig:teaser} (b)). The DeltaTouch device provides a sense of touch at one point in the palm of the user. The unique feature is that display can generate 3D force vector at the contact point to simulate the direction of applied force and sliding direction with arbitrary pressure. The inverted Delta mechanism is proposed to achieve compact and lightweight structure. The vibration at the contact point is achieved by a vibration motor mounted in the end effector and can be used with vibration motors on the fingertips for simulation of the texture of the object. The display is worn on the user's hand in such a way that the bottom platform is fixed on the palm via elastic tapes, while the end effector can freely move along the palm within the boundary of the base platform. And the vibrations motors are fastened to the fingertips via elastic tapes with velcro.
The VR setup includes HTC Vive Pro base stations and head-mounted display (HMD), and Leap Motion controller attached to the headset for tracking the user's hands motion in VR.

\section{VR Applications}
\begin{figure} [h]
  \centering
  \includegraphics[width=1\linewidth]{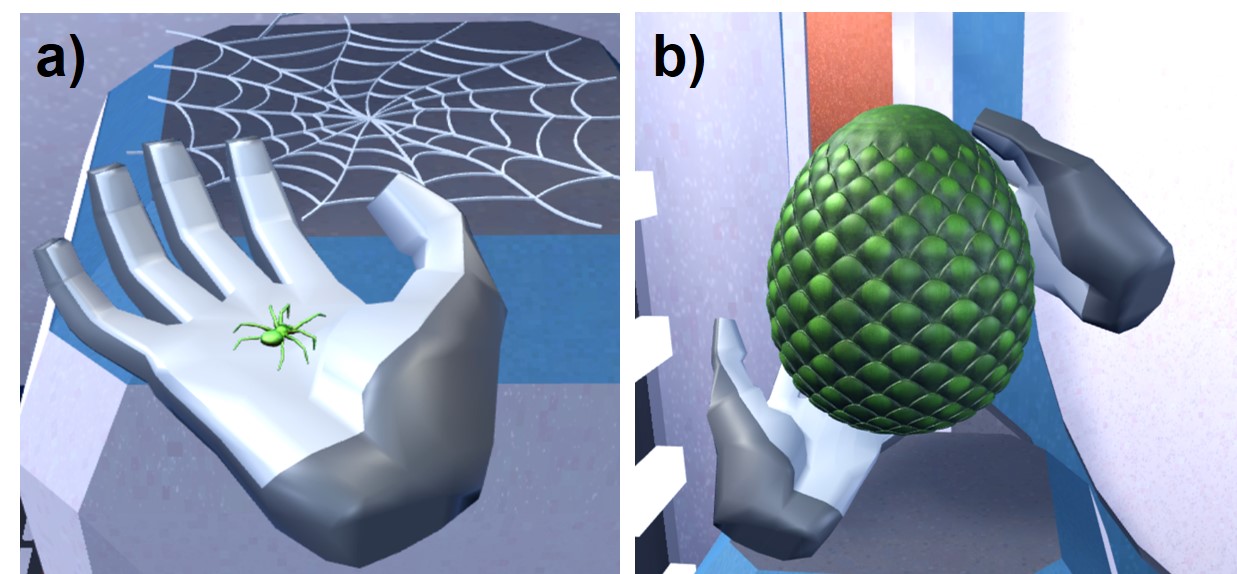}
  \caption{\small Interactions with VR objects: a) AnimalFeel application, b) Interaction with the shaking egg.}
  \label{fig:app1}
\end{figure}

We created several VR applications with different scenarios of user-objects interaction to demonstrate the various types of provided tactile feedback. The applications were implemented with the Unity game engine.
During the demonstration user will wear TouchVR haptic interface in both hands and HTC Vive HMD. The first application is designed to demonstrate interactions with different objects in the VR room. The user can feel the motion of a spider in any direction and its weight on the palm or pulse beat and shake of the dragon egg reproduced by the haptic stimuli on the palm of the user by DeltaTouch display and vibration motors (Fig. \ref{fig:app1}). 
\par
\begin{figure} [h]
  \centering
  \includegraphics[width=0.8\linewidth]{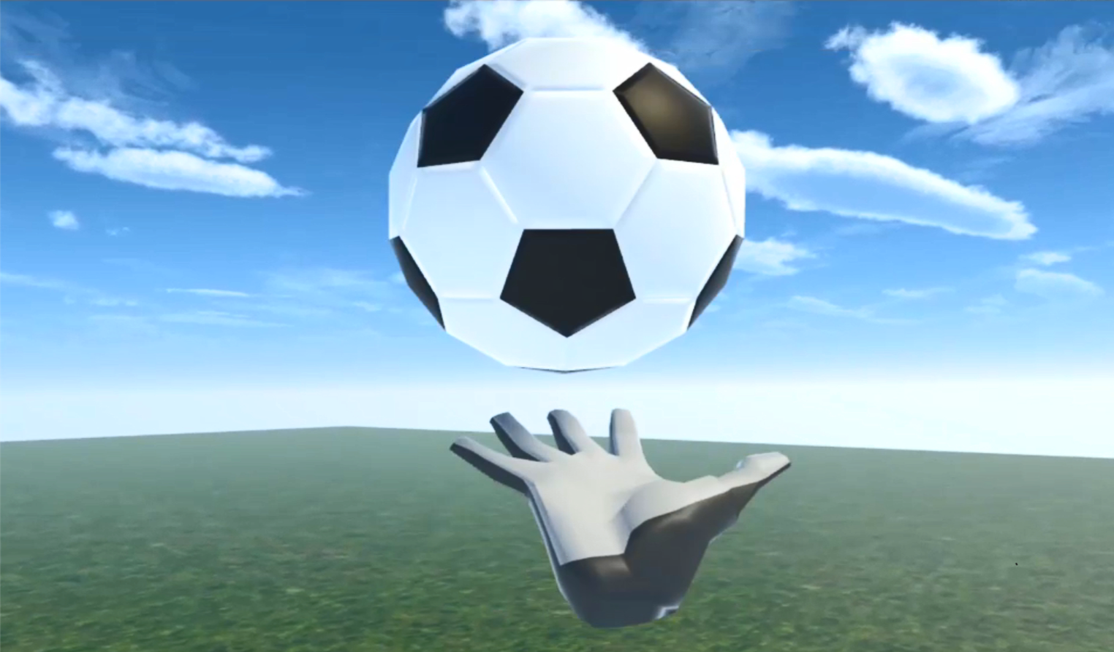}
  \caption{\small  BallFeel application.}
  \label{fig:ball}
\end{figure}

 The application BallFeel is designed to demonstrate the sensation of bouncing the ball in the hands of the user to get the immersive experience of VR games (Fig. \ref{fig:ball}).  In the MatrixTouch application, the user can immerse into the digital matrix, representing a 3D array of running lines (Fig. \ref{fig:teaser} (c)). The RoboX application allows the player to operate the object at a distance via gestures of the hand with the help of tactile feedback. During the collision of the robot with obstacles, the player feels the vibrotactile feedback (Fig. \ref{fig:teaser} (d)).

\section{Conclusions}
We have proposed a Virtual Reality system augmented by a novel haptic interface with multimodal tactile feedback. A wearable haptic display can present 3D force vector at the user palm and multimodal stimuli (contact, pressure, slippage, texture) at the user's palm, while vibration motors provide the vibrotactile feedback to the fingers.  The future work will be aimed at expanding the multimodal feedback. The additional modality will be provided by attaching a Peltier element at the tip of the haptic display. Thus, the user will be capable of experiencing continuous thermal feedback along the palm skin during motion of DeltaTouch. The haptic interface can be applied in medical and rehabilitation systems, teleoperation, and VR simulators (e.g., NurseSim \cite{Nakagawa2014}). The developed wearable cutaneous interface can potentially significantly improve the immersion into VR experience.  


\bibliographystyle{ACM-Reference-Format}
\bibliography{ref}


\end{document}